\pdfoutput=1
\documentclass[11pt]{article}

\usepackage[final]{acl}
\usepackage{times}
\usepackage{latexsym}
\usepackage[T1]{fontenc}
\usepackage[utf8]{inputenc}
\usepackage{microtype}
\usepackage{inconsolata}
\usepackage{graphicx}
\usepackage{booktabs}
\usepackage{longtable}

\title{Multilingual Large Language Models and Curse of Multilinguality}

\author{Daniil Gurgurov\textsuperscript{\normalfont1,2} \quad 
Tanja Bäumel\textsuperscript{\normalfont2} \quad Tatiana Anikina
\textsuperscript{\normalfont2} \\
  \textsuperscript{1}Universität des Saarlandes\\ \textsuperscript{2}German Research Center for Artificial Intelligence (DFKI)\\
  \small{\texttt{\{daniil.gurgurov, tanja.baeumel, tatiana.anikina\}@dfki.de}}
}

\begin{document}

\maketitle

\begin{abstract}
Multilingual Large Language Models (LLMs) have gained large popularity among Natural Language Processing (NLP) researchers and practitioners. These models, trained on huge datasets, show proficiency across various languages and demonstrate effectiveness in numerous downstream tasks. This paper navigates the landscape of multilingual LLMs, providing an introductory overview of their technical aspects. It explains underlying architectures, objective functions, pre-training data sources, and tokenization methods. This work explores the unique features of different model types: encoder-only (mBERT, XLM-R), decoder-only (XGLM, PALM, BLOOM, GPT-3), and encoder-decoder models (mT5, mBART). Additionally, it addresses one of the significant limitations of multilingual LLMs - the curse of multilinguality - and discusses current attempts to overcome it.
\end{abstract}

\section{Introduction}

Large Language Models (LLMs) \cite{devlin2018bert, lewis2019bart, liu2019roberta} have made a significant impact on the field of Natural Language Processing (NLP), showing effectiveness in various tasks. The remarkable aspect of LLMs is their capacity to learn a language during pre-training and enhance their expertise for specific tasks during fine-tuning. Pre-training involves the acquisition of knowledge, where a model grasps language structures by analyzing huge datasets. Fine-tuning, on the other hand, specializes the model by adjusting its parameters so that it can perform specific downstream tasks using a smaller set of examples compared to those used in pre-training.

Another significant advancement involves teaching a model to comprehend multiple languages, leading to the concept of multilingual LLMs \cite{mbert, pires2019multilingual, conneau2019unsupervised, xue2020mt5, liu2020multilingual}. While monolingual LLMs focus on understanding patterns within a single language, multilingual LLMs simultaneously learn from multiple languages. This is accomplished by exposing these models to data from various languages during the pre-training phase. Furthermore, variations in the architectures of multilingual LLMs contribute to their strengths in certain tasks while potentially limiting their effectiveness in others. 

This paper aims to provide a brief overview of the architectures of the most prominent multilingual LLMs, including details such as their pre-training objective functions, data sources, tokenization schemas, the number of languages supported, and the peculiarities of each individual multilingual LLM. Subsequently, the primary challenge facing multilingual LLMs, known as the "curse of multilinguality" \cite{conneau2019unsupervised}, and the current attempts to solve it, are discussed.

\begin{table*}[ht]
    \centering
    \small
    \hspace*{-0.5cm}
    \begin{tabular}{@{}lclcl@{}}
        \toprule
        \textbf{Model} & \textbf{Architecture} & \textbf{Training Data Sources} & \textbf{Languages} & \textbf{Tokenization Schema} \\
        \midrule
        mBERT & Encoder & Wikipedia & 104 & WordPiece \\
        XLM-R & Encoder & CC & 100 & SentencePiece \\
        mBART & Encoder-Decoder & CC25 & 25 & SentencePiece \\
        mT5 & Encoder-Decoder & C4 & 101 & SentencePiece \\
        XGLM & Decoder & CC100-XL & 134 & SentencePiece \\
        PALM & Decoder & Wikipedia, books, webpages, social media \& source code & 124 & SentencePiece \\
        BLOOM & Decoder & ROOTS, OSCAR & 46 & Byte Pair Encoding \\
        GPT-3 & Decoder & CC, Wikipedia, WebText2, Books1\&2 & >95 & Byte Pair Encoding \\
        \bottomrule
    \end{tabular}
    \caption{Comparison of Multilingual Large Language Models}
    \label{tab:multilingual_llms}
\end{table*}

\section{Technicalities}
This section delves into the technical details behind multilingual LLMs, providing a solid foundation for thorough conceptual understanding of their inner workings.

\subsection{Architectures}
LLMs, including multilingual LLMs, are typically constructed using the Transformer architecture \cite{vaswani2017attention}. These models can be categorized into three main architectural types: Encoder-only, Decoder-only, and Encoder-Decoder. 

\begin{figure}[h]
  \centering
  \includegraphics[width=1.0\linewidth]{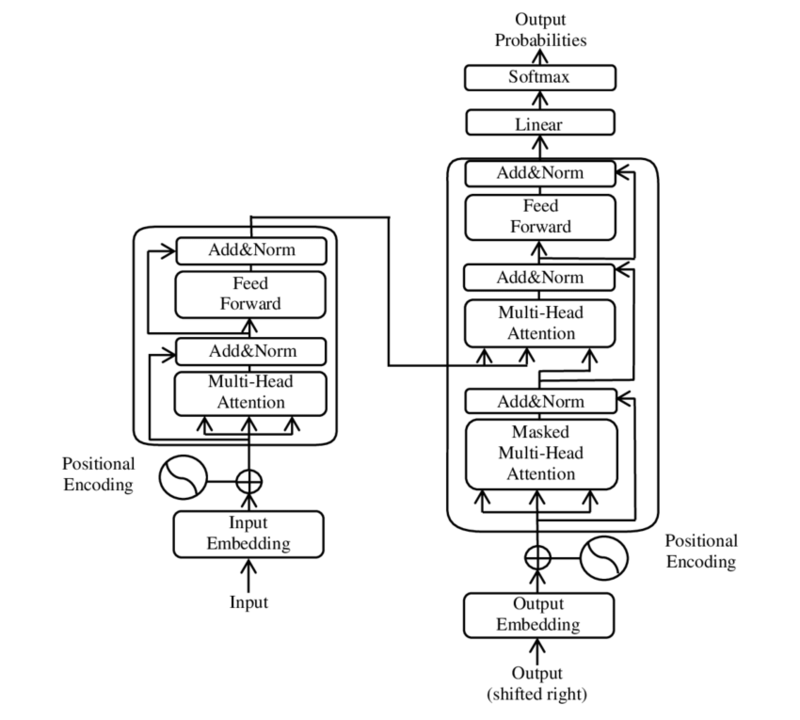}
  \caption{The Transformer Architecture \cite{vaswani2017attention}}
  \label{figure:transformer}
\end{figure}

The original Transformer implementation was intended for machine translation purposes and was of the Encoder-Decoder type, since it included both an Encoder and a Decoder. The architecture is illustrated in figure \ref{figure:transformer} and described as follows. 

The first stage of the architecture involves an Encoder. Initially, the input sentence is split into tokens, which are mapped to their vocabulary indices and converted into embeddings within the "Input embeddings" layer. Simultaneously, positional encoding vectors \cite{gehring2017convolutional} are added to the original token embeddings to capture token position information. The embeddings then undergo a multi-head self-attention mechanism, allowing the model to focus on relevant information while generating the output and enhancing the ability to capture long-range dependencies. This is achieved by calculating attention scores based on the similarity between each token and all other tokens in the sequence. The attention step is followed by the "Add \& Norm" layer \cite{ba2016layer}, serving as a normalization procedure and enhancing gradient flow by combining initial token embeddings to the multi-head attention output. Subsequently, the outputs from the previous step pass through a two-layer Feed Forward Neural Network (FFNN) \cite{ffnn}, followed by another "Add \& Norm" layer. FFNN refines and enriches the token representations coming from the self-attention mechanism, helping the model to capture both local and global contextual information. Finally, the representations are fed into a multi-head self-attention mechanism of the decoder. 

The decoder follows a similar structure to the encoder but with some important differences. The input sequence is tokenized, converted into embeddings, and positionally encoded. The decoder's self-attention is applied in two steps: masked self-attention and subsequent multi-head self-attention. The masked self-attention mechanism ensures that each position in the encoder can only attend to positions before it, which prevents information leakage from future tokens. After the first attention steps, "Add \& norm" is applied, and outputs move to a decoder-encoder attention layer, enabling the encoder to attend to the decoder output. This mechanism facilitates the alignment of input and output sequences, helping with information capture for token generation. Then, "Add \& Norm" is used, and token embeddings pass through another two-layer FFNN, followed by "Add \& Norm". At this stage, the representations for output are generated, going through a final layer and producing a probability distribution over the vocabulary. The model outputs the token with the highest probability for each position. During training, the model is optimized using backpropagation and an optimization algorithm \cite{dreyfus1990artificial}.

\subsubsection{Encoder-Decoder}
In the context of multilingual LLMs, the Encoder-Decoder architecture is utilized for tasks such as machine translation. An exemplary model employing this architecture is the mBART model \cite{liu2020multilingual}. mBART uses the Transformer architecture in an Encoder-Decoder style, where the encoder processes the input text, and the decoder generates the corresponding translated output. This architecture allows the model to effectively capture language-specific details during translation.

\subsubsection{Encoder-only}
Multilingual Encoder-only architectures are designed for tasks that require understanding input sequences without the need for sequence generation. mBERT \cite{mbert} is a prominent example of an Encoder-only model. Trained on a large set of languages, mBERT's encoder is capable of extracting contextualized representations of input. This makes it valuable for various downstream tasks, such as sentiment analysis and named entity recognition, where understanding language context is important.

\subsubsection{Decoder-only}
In specific multilingual cases, a Decoder-only architecture is beneficial. Models like XGLM \cite{lin2021few} follow a Decoder-only configuration. This architecture is particularly useful for tasks where the focus is on generating sequences, such as language modeling or text completion. XGLM is a powerful decoder model that generates coherent sequences in various languages without the need for an encoder.

\subsection{Objective Functions}
In this section, the most popular objective functions used for pre-training multilingual LLMs are described: Masked Language Modeling (MLM) \cite{devlin2018bert}, Causal Language Modeling (CLM) \cite{lample2019cross}, Next Sentence Prediction (NSP) \cite{devlin2018bert}, and Translation Language Modeling (TLM) \cite{lample2019cross}. These pre-training objective functions serve as foundational tasks for training language models.

\subsubsection{MLM}
Inspired by a Cloze task \cite{taylor1953cloze}, Masked Language Modeling involves randomly masking certain words in a sentence and training the model to predict these masked words based on the context provided by the surrounding words. This objective function teaches the model to understand the relationships and dependencies between different words in a sentence, promoting a deeper understanding of syntax, semantics, and contextual nuances within the text.

\subsubsection{CLM}
Causal Language Modeling is designed for autoregressive models, whose focus is on the next token generation. It involves predicting the next word in a sequence given the preceding context. This pre-training objective is particularly effective for tasks where the order of the input sequence is important, as the model learns to capture dependencies and temporal relationships between words.

\subsubsection{NSP}
Next Sentence Prediction trains models to predict whether a given pair of sentences is contiguous or not. The models learn to understand the coherence and logical flow between sentences. This pre-training objective is useful for tasks requiring comprehension of discourse and context, such as question-answering and document summarization. 

\subsubsection{TLM}
Translation Language Modeling extends the pre-training objective to involve parallel data in different languages. The sentences in different languages are concatenated, and words are randomly masked in each of them. Predicting the missing words in this case requires the model to understand the semantic relationships between sentences in different languages and helps in capturing cross-lingual representations. TLM is particularly beneficial for multilingual models, enabling them to transfer knowledge across languages.

\subsection{Pre-training data}

In the pre-training phase, multilingual LLMs leverage two primary types of data: large monolingual corpora in individual languages and parallel corpora across languages. The data for most models is taken from one of the following datasets: Wikipedia \cite{vrandevcic2014wikidata}, Common Crawl (CC) Corpus \cite{wenzek2019ccnet}, ROOTS \cite{laurençon2023bigscience}, OSCAR \cite{oscar}, WebText2 \cite{pile}, Books1, and Books2 \cite{brown2020language}. The choice of training corpora varies among multilingual LLMs. For instance, mBERT leverages Wikipedia as its training data, while XLM-R uses the more extensive CC corpus. 

\subsubsection{Wikipedia}
The Wikipedia dataset includes cleaned articles in various languages, sourced from Wikipedia dumps. Each language has a separate subset, with examples representing entire articles. The content is cleaned by removing markdown and extra sections.

\subsubsection{Common Crawl}
The Common Crawl corpus is a vast collection of web data accumulated over more than 10 years through web crawling. It includes raw web page data, metadata extracts, and text extracts. This non-curated dataset includes web pages in numerous languages, providing a rich source for multilingual pre-training.

\subsubsection{ROOTS}
The Responsible Open-science Open-collaboration Text Sources (ROOTS) corpus is a comprehensive dataset developed by the BigScience workshop, an international and multidisciplinary initiative focusing on researching and LLMs with an emphasis on ethics, harm, and governance. Spanning 1.6 terabytes, the ROOTS corpus serves as a foundational resource for training large-scale language models.

\subsubsection{OSCAR}
The OSCAR project is an open-source initiative aimed at providing web-based multilingual data for ML and AI applications. It offers large quantities of raw data obtained through high-performance data pipelines. OSCAR focuses on improving data quality, particularly for low-resource languages.

\subsubsection{WebText2}
OpenWebText2 is an improved version of the original OpenWebTextCorpus. It covers all Reddit posts from 2005 to April 2020, with additional months becoming available over time. This dataset is another contribution to the diverse training material for multilingual language models.

\subsubsection{Books1 and Books2}
Books1 and Books2 comprise internet-based book corpora, featuring a random sampling of public domain books, as well as modern published literature in e-book format. The contents are drawn from available online books, offering a mix of historical and contemporary literature. It was used for pre-training GPT-3 \cite{brown2020language} and does not seem to be publicly available. 

\subsection{Languages}
Multilingual LLMs demonstrate diversity in terms of the number of languages they support. Models like XLM-R show huge multilingual capabilities, facilitating approximately 100 languages. In contrast, models like mBART target a smaller set of languages, as shown in Table \ref{table:multilingual}. Managing the imbalance in pre-training data among languages, especially when dealing with a large number of them, is a challenge. For example, high-resource languages like English have significantly more data available compared to lower-resource languages like Maltese or Odia \cite{joshi2020state}. To address these imbalances, multilingual LLMs often employ exponentially smoothed weighting \cite{mbert}. This approach ensures relatively fair representation of low-resource languages in the model's training data, preventing them from being underrepresented in the overall vocabulary.

\subsection{Tokenization Techniques}

Tokenization is a critical part in language processing, and several techniques have been developed to represent words effectively. In this section, three widely used tokenization methods are introduced: Byte Pair Encoding (BPE), WordPiece, and SentencePiece.

\subsubsection{Byte Pair Encoding}
BPE \cite{sennrich2016neural} is a subword tokenization technique that builds a vocabulary by iteratively merging the most frequent pairs of consecutive bytes. This method is particularly effective in representing both common words and rare subword units in a wide range of languages and character sets.

\subsubsection{WordPiece}
WordPiece \cite{schuster2012japanese} is another subword tokenization algorithm that starts with a vocabulary of individual characters and iteratively merges the most frequent character pairs. This process creates a vocabulary of subword units, allowing the model to represent words as combinations of subword tokens.

\subsubsection{SentencePiece}
SentencePiece \cite{kudo2018sentencepiece} is a text tokenizer that works at the subword level. Unlike the previous tokenizers, it takes the input sequence as a raw stream, including the space in the collection of characters. It employs a unigram language model to tokenize text into pieces, making it suitable for various languages. It was introduced as a solution for the problem posed by languages that do not use spaces to separate words. 

\section{Multilingual Large Language Models}
This section presents prominent examples of multilingual LLMs categorized into three types: Encoder-Only, Decoder-Only, and Encoder-Decoder, along with their intricacies. Table \ref{tab:multilingual_llms} summarizes the important points regarding all presented models.

\subsection{Encoder-Only Models}
\subsection{mBERT}
The first notable example of a genuinely multilingual neural LLM is the multilingual Bidirectional Encoder Representations from Transformers (mBERT) \cite{devlin2018bert, wenzek2019ccnet}. This model adapts the architecture similar to the original BERT \cite{devlin2018bert}, consisting of 12 Transformer layers and operating as an encoder-only model. The distinguishing factor lies in the model being trained on the entire Wikipedia data covering 104 languages \cite{pires2019multilingual}, in contrast to the original BERT, which was trained only on English. Additionally, mBERT employs a shared vocabulary for all languages based on the WordPiece tokenization schema. Pre-training mBERT involves two unsupervised tasks: MLM and NSP. 

\subsection{XLM-R}
Cross-lingual Language Modeling RoBERTa (XLM-R) \cite{conneau2019unsupervised} is a 12-layers Transformer-based multilingual model that adopts an encoder-only architecture following the XLM \cite{lample2019cross} approach. It integrates RoBERTa, introducing simple improvements to the learning procedure. Unlike the original XLM, which employed both MLM and TLM objectives for pre-training, XLM-R primarily utilizes the MLM objective. XLM-R is available in two configurations - XLM-R Base (270M parameters) and XLM-R XL (550M parameters). Both variants are trained on the clean CommonCrawl corpus covering 100 languages. The tokenization method utilized by the model is SentencePiece.

\subsection{Decoder-Only Models}
\subsubsection{XGLM}
XGLM \cite{XGLM}, or Cross-lingual Generative Language Model, is a family of multilingual models trained on the extensive CC100-XL dataset, this is CommonCrawl snapshots including data from 2013 to 2020. The models vary in size and configuration, with parameters ranging from 564 million to 7.5 billion, language coverage from 30 to 134, and number of layers from 24 to 48. They utilize the SentencePiece tokenizer and focus on CLM during pre-training, aiming to predict the next token given the previous ones. The largest XGLM model, with 7.5 billion parameters, achieved state-of-the-art performance in few-shot learning across multiple languages at the time of the release, surpassing GPT-3 in tasks like commonsense reasoning and natural language inference.

\subsubsection{PALM}
PaLM \cite{chowdhery2022palm}, or Pathways Language Model, is a multilingual model trained on a diverse corpus consisting of content from Wikipedia, books, web pages, social media, and source code. While exact details on the training data are not provided, it is stated that PaLM is trained on 124 languages using the SentencePiece tokenizer. Pre-training is done using Pathways, a system enabling highly efficient pre-training across multiple TPUs. The objective function used for pre-training is autoregressive language modeling (CLM). The model comes in a few configurations: 8B parameters and 32 Transformer layers, 62B parameters and 64 Transformer layers, and 540B parameters and 118 transformer layers. The largest variant, PaLM 540B, achieves competitive results across various tasks, despite being trained on approximately 22\% of non-English data out of the total 780 billion training tokens.

\subsubsection{BLOOM}
BLOOM \cite{workshop2023bloom}, standing for Bloom Language Model, is a significant step towards open-sourcing LLMs. With a capacity of 176 billion parameters and 70 Transformer layers, BLOOM is an open-access LLM developed through collaboration among hundreds of researchers. BLOOM is a decoder-only Transformer language model trained on the ROOTS corpus, which consists of data from hundreds of sources, including OSCAR, covering 46 natural languages and 13 programming languages. Similar to other models, BLOOM utilizes CLM during training and uses BPE tokenization. Despite being open-source, BLOOM demonstrates competitive performance across a wide range of benchmarks. Moreover, its performance can be further enhanced through multitask prompted fine-tuning. To support future research and applications using LLMs, the models and code for BLOOM are publicly released, which promotes transparency and collaboration in the field of NLP.

\subsubsection{GPT-3}
GPT-3 \cite{brown2020language}, short for Generative Pre-trained Transformer 3, is a widely known LLM developed by OpenAI. It is a decoder-only Transformer architecture, consisting of 175 billion parameters and 96 Transformer layers. GPT-3 is trained using the CLM task on a diverse dataset covering Common Crawl (CC), Wikipedia, WebText2, and Books1 \& 2. These sources provide a wide range of textual data, enabling GPT-3 to understand and generate human-like text across various domains. The model covers over 95 languages, which is not explicitly stated in the original report and estimated by users. For tokenization, GPT-3 employs BPE. Additionally, GPT-3 incorporates advanced techniques such as few-shot learning, enabling it to perform new tasks with minimal instructions. Despite its impressive performance and wide adoption, GPT-3 is not openly accessible for training or fine-tuning by external researchers. However, its capabilities have captured significant interest, serving as a benchmark for the development of new language models.

\subsection{Encoder-Decoder Models}

\subsection{mBART}
The next model is the multilingual Bidirectional Autoregressive Transformer (mBART) - a multilingual sequence-to-sequence (Seq2Seq) denoising model \cite{liu2020multilingual}. It stands out as one of the pioneering models employing both an encoder and a decoder. The model is trained by applying the BART objectives \cite{lewis2019bart} on extensive monolingual corpora across a variety of languages. This Seq2Seq model is pre-trained by denoising full texts in various languages, and it is primarily intended for translation tasks. During pre-training, the model tackles noise in the input by masking phrases and introducing sentence order permutations, which is a variant of MLM. Consequently, it learns to reconstruct the masked words based on context, developing an understanding of word relationships, and to correctly order sentences, thereby capturing relationships between sentences. A single Transformer model with 12 encoder and 12 decoder layers consisting of 680 million parameters is used to execute training. The pre-training data is the CC25 dataset extracted from the Common Crawl, which comprises of 25 languages and is tokenized using the SentencePiece schema. 

\subsection{mT5}
The Multilingual Text-to-text Transfer Transformer, mT5 \cite{xue2020mt5}, is the model that leverages the T5 architecture \cite{raffel2020exploring}. Its training data comes from Multilingual C4 (mC4) dataset, comprising data in 101 languages sampled from Common Crawl and tokenized using the SentencePiece schema. An advantageous feature of the T5 architecture lies in its unified "Seq2seq" format, where the model generates text conditioned on given text inputs. Employing this "text-to-text" structure, mT5 uses a standard Encoder-Decoder Transformer architecture. For pre-training, a variation of MLM, involving the replacement of consecutive spans of input text with a masked token and predicting them, is used. Notably, mT5 exists in several variants, distinguished by the quantity of data used for training and the number of parameters in the model. These variants include mT5-Small, mT5-Base, mT5-Large, mT5-XL, and mT5-XXL. The number of parameters ranges from 300 million to 13 billion and number of layers - from 12 to 24 in both encoder and decoder.

\section{Curse of Multilinguality}
This chapter addresses one of the biggest challenges faced by multilingual LLMs - curse of multilinguality - and explores its potential solutions to overcome it.

\subsection{Motivation}
The curse of multilinguality was introduced by \citeauthor{conneau2019unsupervised} and refers to the challenges and limitations that arise when developing multilingual LLMs \cite{conneau2019unsupervised}. As the number of languages increases, a transfer-dilution trade-off occurs, diminishing the per-language capacity and consequently impacting model performance. Initially, adding similar higher-resource languages during pre-training can enhance the performance of low-resource languages. However, beyond a certain point, the curse of multilinguality starts to manifest, leading to decreased performance across all languages \cite{aharoni-etal-2019-massively}. This emphasises the complexity of achieving universality in multilingual models and points at the need to address key factors contributing to the curse of multilinguality.

\subsubsection{Linguistic Diversity}
Languages show significant diversity in grammar, syntax, vocabulary, and cultural nuances, posing a significant challenge for multilingual model development \cite{conneau2019unsupervised}. Accommodating such diversity while maintaining model efficiency and performance becomes increasingly challenging with multiple languages, worsening the effects of the curse of multilinguality.

\subsubsection{Data Sparsity}
The availability of training data varies widely across languages, with some languages possessing a lot of resources while others being severely under-resourced \cite{joshi2020state}. The impact of data sparsity on training efficient multilingual models is large, as highlighted by \citeauthor{conneau2019unsupervised} \cite{conneau2019unsupervised}. Models may struggle to generalize well to languages with limited training data, further complicating the curse of multilinguality.

\subsubsection{Model Complexity and Scale}
Multilingual models require increased complexity and scale to accommodate the diverse linguistic characteristics of multiple languages \cite{conneau2019unsupervised}. Balancing model size and computational resources is crucial, as overly complex models may become impractical for real-world deployment. \citeauthor{conneau2019unsupervised} emphasize the importance of addressing these challenges to mitigate the curse of multilinguality.

\subsection{Potential Solutions}

Recent studies have pointed at reduced monolingual and cross-lingual capabilities of models due to curse of multilinguality, particularly for low-resource languages not having a lot of pre-training data \cite{wu-dredze-2020-languages, lauscher-etal-2020-zero, artetxe-etal-2020-cross}. Various approaches have been proposed to address this problem and improve model performance.

\subsubsection{Modular Multilingual Architecture}
\citeauthor{pfeiffer-etal-2022-lifting} propose X-MOD, a modular multilingual architecture that combines shared and language-specific parameters as a way of uplifting curse of multilinguality \cite{pfeiffer-etal-2022-lifting}. 
X-MOD initializes modular models during pre-training, facilitating inexpensive expansion to new languages afterwards.

\paragraph{Architecture:} X-MOD is an extension of the transformer-based architecture from mBERT and XLM-R, incorporating language-specific modules at every transformer layer. Each language has its own module, while attention and feed-forward components are shared, facilitating efficient training and inference without significantly increasing computational costs.

\paragraph{Pre-training:} X-MOD is pre-trained using MLM on combined monolingual corpora in multiple languages. Efficient utilization of language-specific modules ensures effective handling of linguistic diversity during pre-training.

\paragraph{Extending to New Languages:} The modular design allows X-MOD to be extended to new languages after pre-training, with minimal impact on performance in pre-trained languages. This adaptability is achieved through the learning of new embeddings and adapter modules for the target language via MLM.

\paragraph{Fine-tuning on Downstream Tasks:} X-MOD can be fine-tuned for cross-lingual downstream tasks by selectively updating shared weights on source language data while keeping modular components frozen, ensuring efficient adaptation to target languages.

\subsubsection{Alternative Approaches}
Several alternative approaches have been proposed to extend multilingual and monolingual LLMs to other languages and improve their cross-lingual capabilities as a mitigation of curse of dimensionality.

\paragraph{Training New Embedding Layer:} \citeauthor{artetxe-etal-2020-cross} propose training a new embedding layer with a corresponding target-language tokenizer to extend monolingual models to new languages \cite{artetxe-etal-2020-cross}. This approach helps with language extension while maintaining model stability.

\paragraph{Transliteration and Subword Mappings:} Transliteration-based approaches and subword mappings offer potential solutions for incorporating additional languages into multilingual models \cite{muller-etal-2021-unseen, vernikos-popescu-belis-2021-subword-mapping}. These methods contribute to the expansion of multilingual model capabilities.

\paragraph{Adapter-based Approaches:} Adapter-based approaches, proposed by \citeauthor{pfeiffer-etal-2020-mad}, offer efficient solutions for adapting multilingual LLMs to specific languages or extending multilingual them to unseen languages and overcoming curse of dimensionality \cite{pfeiffer-etal-2020-mad}. While achieving significant performance gains, these approaches may build upon sub-optimal parameter initializations.

\section{Conclusion}
In summary, multilingual LLMs stand out as robust tools in NLP, showcasing proficiency across multiple languages and tasks. This paper has provided insights into key models such as mBERT, XLM-R, mBART, mT5, XGLM, PALM, BLOOM, and GPT-3, highlighting their underlying architectures and technical characteristics. These models represent a significant advancement in NLP, having deep multilingual understanding and potential for cross-lingual applications. Their development plays a great role in advancing language technology and preserving linguistic diversity across the world. 

Despite their capabilities, multilingual LLMs face challenges such as the curse of multilinguality, which constrains their performance across different languages. However, various approaches have been developed and are being pursued to mitigate these challenges. These efforts aim to improve the effectiveness and adaptability of multilingual LLMs.

\bibliography{custom.bib}

\end{document}